\documentclass[runningheads]{llncs}
\usepackage[T1]{fontenc}

\usepackage{microtype}
\usepackage{graphicx}
\usepackage{subfigure}
\usepackage{booktabs} 
\usepackage{amsmath}
\usepackage{hyperref}

\usepackage{tikz}
\usepackage{pgfplots}

\begin{document}

\title{Empirical Capacity Model for Self-Attention Neural Networks}

\author{Aki H\"arm\"a \orcidID{0000-0002-2966-3305} \and
Marcin Pietrasik \orcidID{0000-0001-7559-8658} \and
Anna Wilbik \orcidID{0000-0002-1989-0301}}
\authorrunning{A. Härmä et al.}
%
\institute{Department of Advanced Computing Sciences (DACS), Faculty of Science and Engineering, Maastricht University, The Netherlands \\
\email{aki.harma@maastrichtuniversity.nl}}

\maketitle  

\begin{abstract}
Large pretrained self-attention neural networks, or {\it transformers}, have been very successful in various tasks recently. The performance of a model on a given task depends on its ability to memorize and generalize the training data. Large transformer models, which may have billions of parameters, in theory, have a huge capacity to memorize content. However, the current algorithms for the optimization fall short of the theoretical capacity, and the capacity is also highly dependent on the content. In this paper, we focus on the memory capacity of these models obtained using common training algorithms and synthetic training data. Based on the results, we derive an empirical capacity model (ECM) for a generic transformer. The ECM can be used to design task-specific transformer models with an optimal number of parameters in cases where the target memorization capability of the task can be defined.   
\end{abstract}

\section{Introduction}
\label{intro}
The central processing element of the transformer model \cite{vaswani_attention_2017} is the self-attention circuit \cite{bahdanau_neural_2015-1}, which computes weighted sums of input vectors based on content. Large transformer models, which may have billions of parameters, typically have multiple layers of parallel sets, or {\it multihead}, self-attention circuits, and several other processing units. The parameters are optimized using stochastic gradient backpropagation methods \cite{kingma_adam_2015}. Neural networks based on transformer architecture have been very successful in recent years in various tasks such as natural language processing \cite{brown_language_2020,liang_holistic_2022}, speech recognition \cite{radford_robust_2022}, and image processing \cite{liu_survey_2023}. 

The memory capacity discussed in this paper does not refer to the ability of a model to learn a causal pattern in the sequences, which has been studied, for example, in \cite{xie_explanation_2022,nichani_how_2024-1}. 
With a sufficient number of parameters, a neural network can memorize, or {\it shatter} \cite{vapnik_uniform_1971}, the machine learning problem. The self-attention circuit can be considered as an associative memory that has a certain capacity determined by the number of parameters. Transformer models have a known relation to Hopfield networks \cite{ramsauer_hopfield_2021} and associative memories \cite{radhakrishnan_overparameterized_2020,bietti_birth_2023,schaeffer_bridging_2024}. In theory, transformer models may have a large storage capacity that depends on the architecture choices \cite{kim_provable_2022,mahdavi_memorization_2024}. However, translating those results into actual attainable capacity is difficult. Allen-Zhu {\it et al.} have recently published a series of preprints about the performance of transformer models in using synthetic data derived from a knowledge base \cite{allen-zhu_physics_2024-1,allen-zhu_physics_2024-2,allen-zhu_physics_2024}. The capacity in those works were not defined by exact memorization, which is the scope of the current paper, but the ability to retrieve “one piece of human knowledge” \cite{allen-zhu_physics_2024-2}, which is naturally a meaningful goal but difficult to define in a more general case than synthetic content. Interestingly, they concluded that a transformer model can store 2 bits of knowledge per parameter \cite{allen-zhu_physics_2024-2}. 

The {\sl scaling laws} for large transformer models have been discussed, based on the analysis of training loss, for example, in \cite{kaplan_scaling_2020}, but those results do not directly translate to capacity estimates. Kim {\it et al.} showed that the capacity of transformer models follows a trend similar to an asymptotic theoretical model \cite{kim_provable_2022}. Mahdavi {\it at al.} proposed a model for the capacity for a single self-attention layer which was also compared to numerical performance \cite{mahdavi_memorization_2024}. However, there seem to be few previous studies that address the actual attainable storage capacity of complete transformer models from the perspective of deriving practical design rules.  

In this paper, we measure the capacity of transformer models computationally by training models of different sizes using synthetic data. Based on various experiments with different sizes and architecture parameters, we fit a function that gives the expected capacity of the model. The function outperforms a polynomial function at a fraction of the trainable parameters. The prediction of the model can be considered as an empirical lower bound for the capacity, which can make it a useful tool for designing architectures that meet the required performance criteria of a particular application. We demonstrate that the model can be used to predict the capacity of unseen model architectures. The use of the function can lead to savings in processing costs and also facilitate the development of new Retrieval Augmented Generation, RAG, methods \cite{lewis_retrieval-augmented_2020} based on the memorization capacity of large models.

\section{Capacity in networks}
The basic quality measure for the Hopfield networks is the storage capacity, understood as the maximum number of patterns that are remembered as fixed points by the network. The storage capacity of a Hopfield network with $N$ nodes is $CN/log(N)$ where $C < 1/2$ \cite{mceliece_capacity_1987}, and the stored {\sl memories} are binary vectors of length $N$. It is interesting to note that a Hopfield network of the size of GPT-3 \cite{brown_language_2020}, with 175B parameters, could theoretically have the capacity to memorize 3.4B sequences of 2048 tokens. The reported number of GPT-3 training data tokens is 500B, that is, about 7 percent of the theoretical capacity of the network of this thought experiment. Based on the rule of 2 bits per parameter in \cite{allen-zhu_physics_2024-1} this corresponds to 44 Gb of memorized {\it knowledge}.   

The storage can be measured as mean squared errors (MSEs) between the original and retrieved patterns across different numbers of patterns the model was trained to memorize~\cite{tang2023recurrent}.
Some authors use recall rate~\cite{ZHONG20161154}, measured as a rate of how many patterns can be reconstructed correctly given a pattern with a certain number of missing bits.
Steinberg {\it et al.} \cite{steinberg2022associative} found that there is a minimum initial cue (relative to the total length) that gives a very small recall error. 
This minimum initial cue is defined as 
\begin{equation*}
    l_c = \frac{L_0}{L},
\end{equation*} 
where $L_0$ is the cue length and $L$ is the pattern length. 

Besides storage capacity, many authors were also interested in measuring the so-called basin of attraction~\cite{steinberg2022associative,tavan1990self}. The basin of attraction is understood as a region around a pattern in which all states are attracted to this pattern within a prescribed time. There is no analytical way to study the basin of attraction except by drawing a graph which shows the number of successes out of many trials against the number of input errors.

An interesting characteristic of Hopfield network is a degree of symmetry, introduced by Krauth~\cite{krauth} and calculated as
\begin{equation} 
 \frac {\Sigma_{i=1}^n \Sigma_{j=1}^n w_{ij} w{ji}}{\Sigma_{i=1}^n \Sigma_{j=1}^n w_{ij}^2}
\end{equation} 
 Many researchers believed that the maximal number of connections and 100\% symmetry of the weight matrix disable to achieve its maximal storage capacity. Others claim that due to the asymmetry of the weights matrix, the number of false attractors increases.

Baum {\it et al.} \cite{baum_capabilities_1988} have shown that a feedforward net of one hidden layer with $N/d$ internal units can perfectly learn an arbitrary two-class classification problem, a {\it dichotomy}, of $N$ $d$ dimensional vectors \cite{baum_capabilities_1988}. The storage capacity of RELU feedforward networks was studied by Vardi {\it et al.} \cite{vardi_optimal_2021}. In their model, the capacity in the big-\cal{O} notation is $\tilde{O}(\sqrt{N})$. \cite{kim_provable_2022} \cite{kim_provable_2022} recently proposed that in a transformer architecture, $\tilde{O}(d + n + \sqrt{nN})$ parameters are enough to memorize $N$ sequences of length $n$ with a token vector dimension $d$. However, this does not seem to translate into a practical design rule that gives a memorization requirement. Mahdavi {et al.} proposed that a single multihead self-attention layer of $\Theta(Hd^2)$ parameters can memorize $O(HN)$ sequences of length $N$. However, their model was based on $d=d_h$ which is not a typical choice in transformer models. 

\section{Transformer models}
The input ${\bf X}$ into the model is a sequence of $N$ discrete symbols $t_i$, or tokens in the NLP terminology. In the model, each sequence of tokens is represented by the vectors of $B$ elements $x_i, i=0,\cdots,N-1$.   
The core component is the self-attention circuit which in the conventional form is given by 
\begin{equation}
    {\rm Attn}(Q, K, V) = {\rm softmax}\left(\frac{QK^T}{\sqrt{d_k}}\right)V
\end{equation}\label{eq:attention}
where $Q=XW_Q$, $K=XW_K$, and $V=XW_V$, 
and matrices $W_Q$, $W_K$, and $W_V$ are $B\times d_h$ matrices of a selected $x_i$ token vector size $B$, and a head dimension $d_h$. The coefficients of the three matrices are trained in the model optimization process. For example, the model of \cite{mahdavi_memorization_2024} suggests that the capacity to memorize $C$ independent data sets is related to $O(HN)$, where $H$ is the number of attention heads.

The self-attention circuit is the key component of the transformer architecture and is mainly responsible for the memorization of training data. The $QK^T$ matrix operation can be rewritten as 
\begin{multline}
        QK^T=XW_Q(XW_K)^T \\=XW_QW_K^TX^T =XW_AX^T,\label{eq:decomp}
\end{multline}
that is, the rightmost {\it quadratic form} with one $d\times d$ matrix $W_A$ is equivalent with the original $QK^T$ key-query form that has two $d_h\times d$ matrices. Usually in LLMs $d>>4d_h$, and therefore the second form of (\ref{eq:decomp}) is more efficient than the last form with a matrix $d\times d$. However, if $d\leq2d_h$, the effective number of parameters in a single attention circuit is $d^2+dd_h\leq3d^2$. 

Moreover, omitting $W_V$, which is commonly done in some transformer implementations, we may write the 
\begin{equation}
    {\rm Attn}(Q, K, V) = MX    
\end{equation}\label{eq:new_attention}
where the matrix $M=f(X)$, where $f()$ is some function learned by the transformer network. This operation of multiplying the array of input vectors in $X$ by a matrix that depends on $X$, based on learning, can be seen as the key operation in transformer models.  

\begin{figure}
    \centering
    \includegraphics[width=4cm]{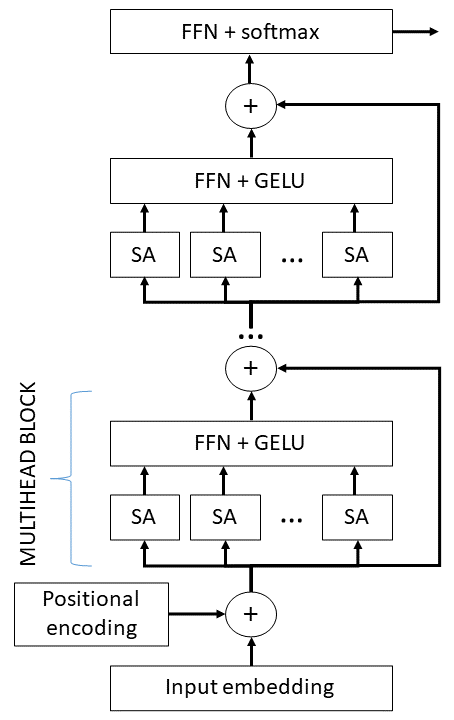}
    \caption{A self-attention model with multiple layers and multiple self-attention heads.}
    \label{fig:model}
\end{figure}

A complete transformer system, see Fig.~\ref{fig:model} consists of multihead attention modules, which have $H$ parallel and independent self-attention (SA) circuits. An array of $H$ SA blocks in parallel on the same input is called a multi-head attention block. Often, large models consist of several layers of multi-head blocks. The normalization and dropout models have been left out of the figure, but they are also a part of the implementation used in the following experiments.

\begin{figure}
    \centering
    \includegraphics[width=8cm]{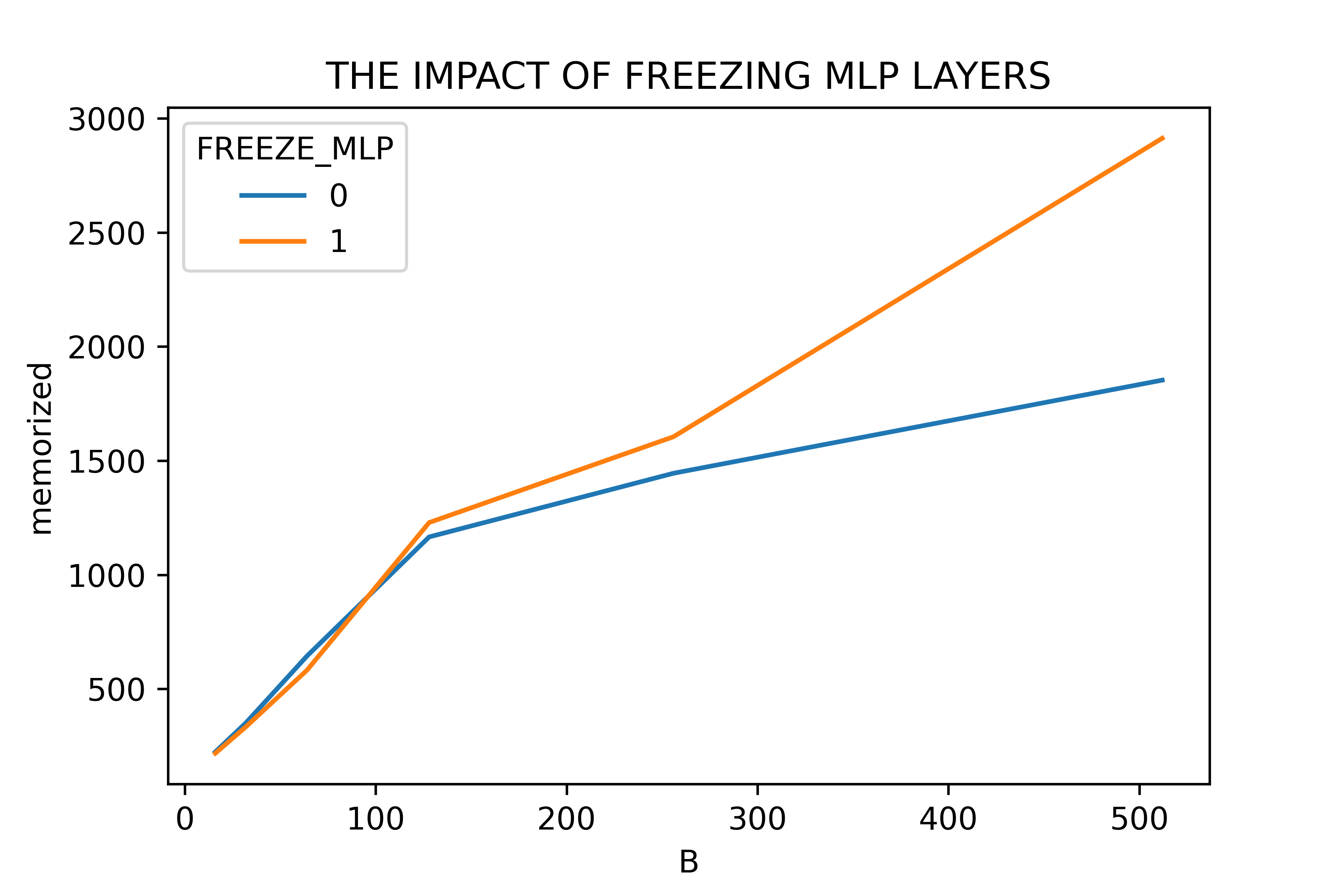}
    \caption{A self-attention model with trained (0) and frozen (1) MLP layers.}
    \label{fig:freezemlp}
\end{figure}

After each layer, there is a dense feedforward (FFN) layer with GELU activation function. One may suggest that most of the memorization capacity of the model is in the SA circuits. To show this, models were trained with trainable FF layers and by freezing the coefficients of the FF layer to an identity operation that maps the inputs to the output without modification. The results of a two-head, two-layer transformer for different sizes of the token vector $B$ are shown in Fig. \ref{fig:freezemlp}. For small vector sizes, the curves are identical, which confirms that the FF layers do not contribute to the memorization of the synthetic content used in this paper. The capacity at the larger values of $B$ seems to even increase after freezing the FFN parameters. It is likely that the fully trainable model would require more iterations to reach the same capacity as the simpler model where the FF layer is replaced by an identity operation.   

\begin{figure}
    \centering
    \includegraphics[width=8cm]{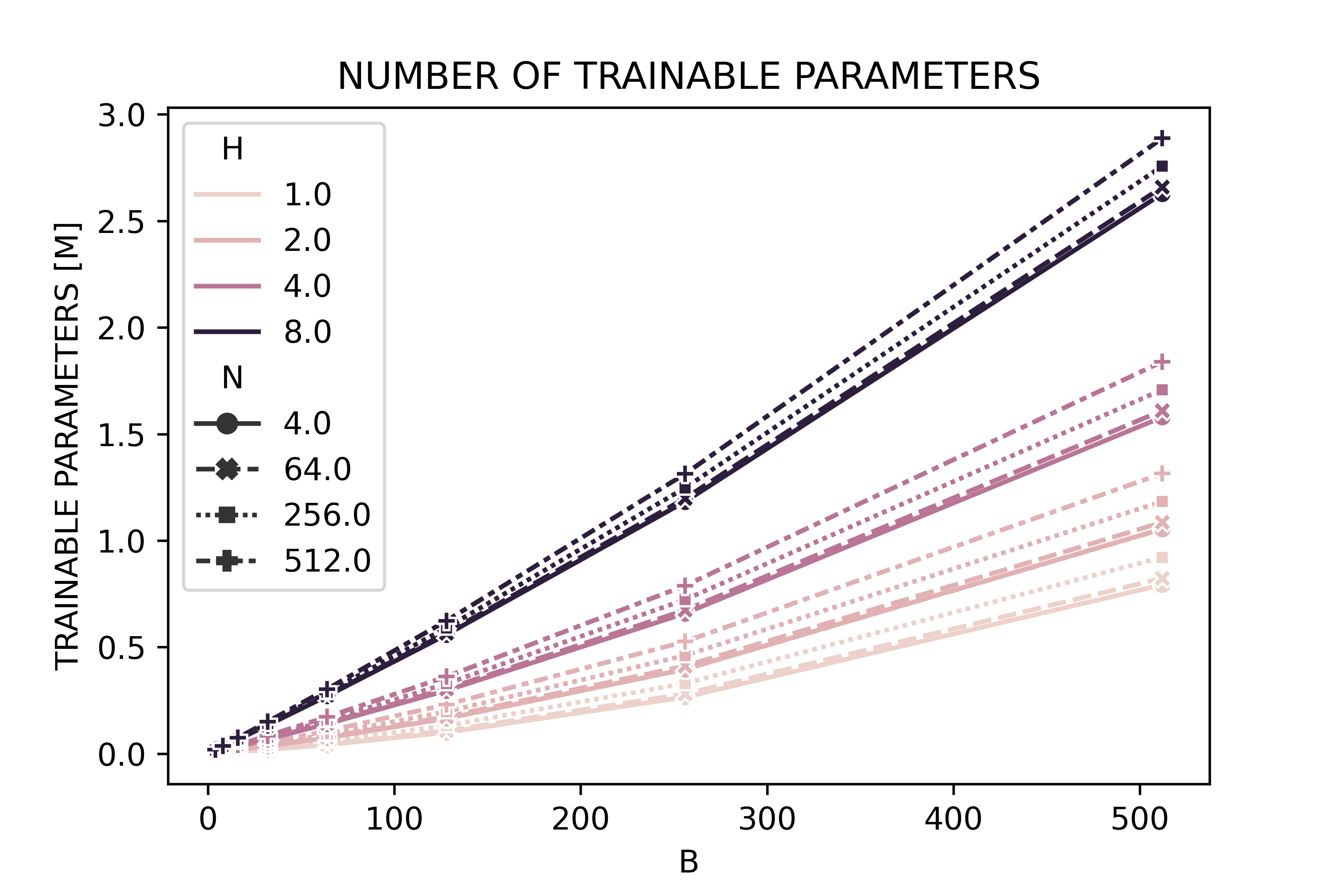}
    \caption{Total numbers of trainable parameters (millions) for different variants of the transformer network.}
    \label{fig:params}
\end{figure}

The total number of parameters of a self-attention network depends on the sizes of the vectors $d_h$ and $d$, the maximum length of the input sequence $N$, and the number of heads $H$ and layers $L$. Fig. \ref{fig:params} shows the total number of {\it trainable parameters} for different configurations. 
In the experiments reported in this paper, we freeze the coefficients corresponding to the input embedding and the final linear model at the end. The trainable parameters are therefore those of the SA and FFN modules.   

\section{Data sequences}
Let us consider the system of $T$ tokens in an autoregressive task of predicting the next token
\begin{equation}
    t_p = {\rm argmax_k}F(t_k| t_i, i=0,\cdots, k-1) 
\end{equation}
where the model $F()$ is a transformer followed by a softmax layer
\begin{equation}
    F(x_i,i\neq k) = {\rm softmax}({\rm Attn}(Q, K, V)H)
\end{equation}
and $H$ is an $T\times N$ matrix with fixed coefficients. 

The modeling task is an autoregressive task to predict the next token from the previous $N-1$ tokens. In this paper, we define the empirical capacity measure $C$ of a network as the number of sequences in which the network can correctly predict the $N^{th}$ token from a sequence of preceding tokens $N-1$. In a system of $T$ different tokens, the probability of {\it guessing} the correct token is $p=1/T$ and the probability that a system gives the guess correctly $r \leq R$ times in a system of $K$ sequences is 
\begin{equation}
P(r<R) = \sum_{r=0}^{R-1} \binom{K}{r} \frac{1}{T}^r (1-\frac{1}{T})^{K-r}\label{eq:bound}
\end{equation}
For example, with $T=128$ and $K=2048$ the probability of $25$ correct guesses (of an untrained network) is $P(r<25) = 0.96$. Furthermore, if the probability of more than, say, 40 right guesses in this system is very unlikely, that is, $1-P(r<40) \approx 2.8\times 10^{-7}$. We may use Equation \ref{eq:bound} to assess the validity of the empirical capacity measures discussed in this paper. Furthermore, for a library size $K$, the expectation for the number of correct guesses is:
\begin{equation}
    c_{\rm offst} = \frac{K}{T} \label{eq:offset}. 
\end{equation}

% ...


\section{Experiments}
There are several ways to address capacity measurement in learning experiments. In this paper, we consider two approaches. In the Maximum Library Size (MLS) method, the goal is that the network memorizes every item of the input vector library, and the measurement is performed by measuring how large the library can be fully memorized. In the Maximum Attainable Capacity (MAC) method, the model is trained with a large library, and the goal is to measure the maximum number of samples the network can memorize. The MLS method is obviously more tedious than the MAC method. 

\subsection{Comparison of the MAC and MLS methods}
\begin{figure}[t]
    \centering
    \includegraphics[width=8cm]{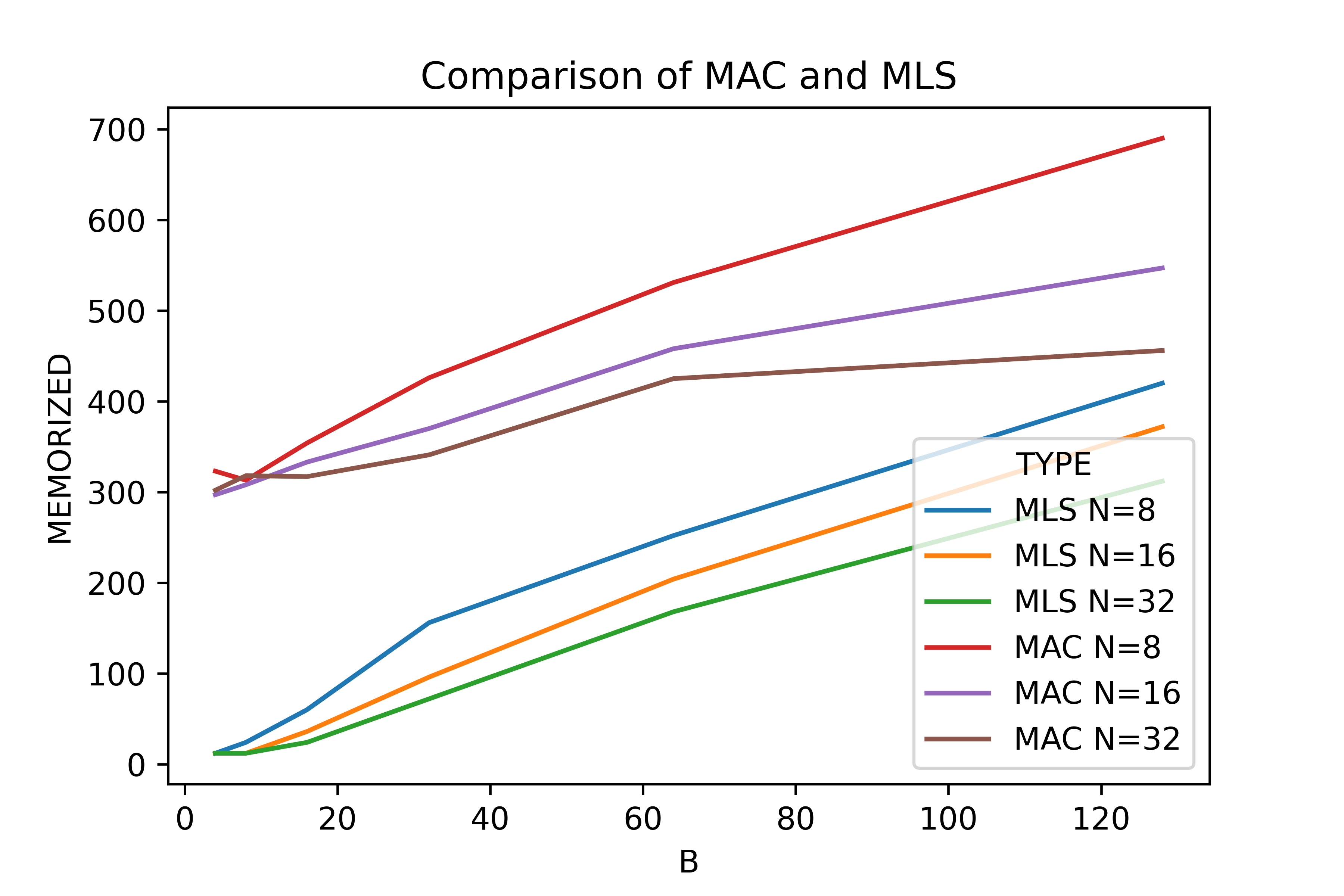}
    \caption{Comparison of the capacity measurements with $H=4, L=1$ in MSL and MAC methods}
    \label{fig:MAC}
\end{figure}

As shown in Fig. \ref{fig:MAC} the MLS and MAC capacity measurements have a similar overall shape but there is an offset in the MAC values. The offset is due to the correct guesses in the MAC method. In this case, the size of the library $K=32000$ and with the token count of $T=128$ we get, by (\ref{eq:offset}), $K/T=250$, which agrees with the offset. 

For larger vector sizes $B$, the MAC capacity, after compensating the offset, seems to be lagging behind the MLS capacity estimates. Therefore, we may consider MLS to represent a lower bound for storage capacity. Since the MAC case represents a more realistic use case, in this paper, we focus on modeling the results of MAC experiments.  

\begin{figure}[t]
    \centering
    \includegraphics[width=7cm]{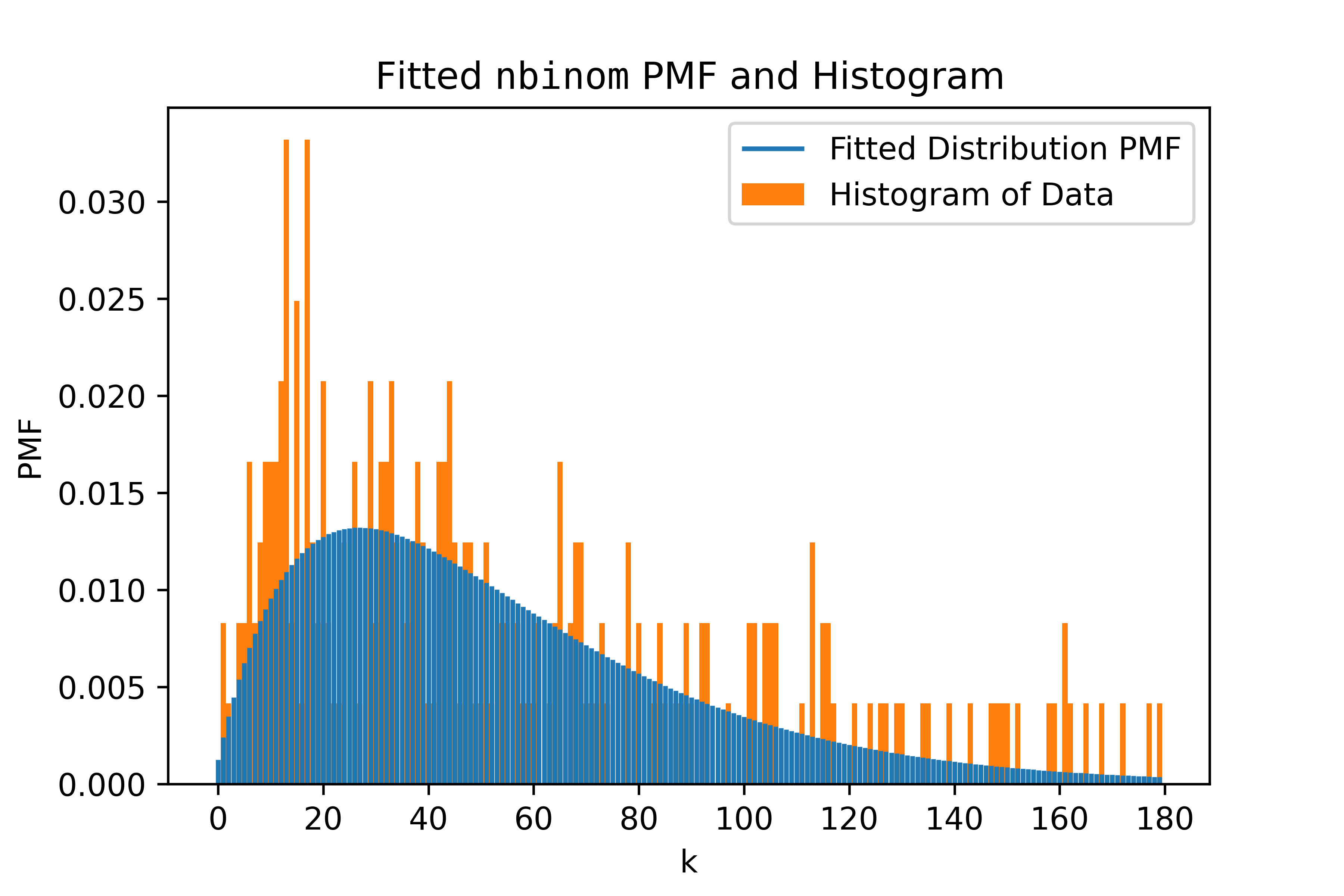}
    \caption{Number of epochs needed to shutter the entire data set}
    \label{fig:negbinom}
\end{figure}

The probability that the model training takes $k$ epochs before the model shutters (MLS condition) can be expected to follow the negative binomial distribution
\begin{equation}
P_r(X=k)={\binom {k+r-1}{k}}(1-p)^{k}p^{r}
\end{equation}
For example, Fig. \ref{fig:negbinom} shows the histogram of the number of epochs for a transformer model to {\it shutter}, that is, learn perfectly all sequences in 1000 independent trials in case of 16 sequences of length $N=8$ and with a token vector size $d=16$, and therefore $2\times d^2=512$ parameters in total. 

\subsection{Impact of batch size}
\begin{figure}[t]
    \centering
    \includegraphics[width=7cm]{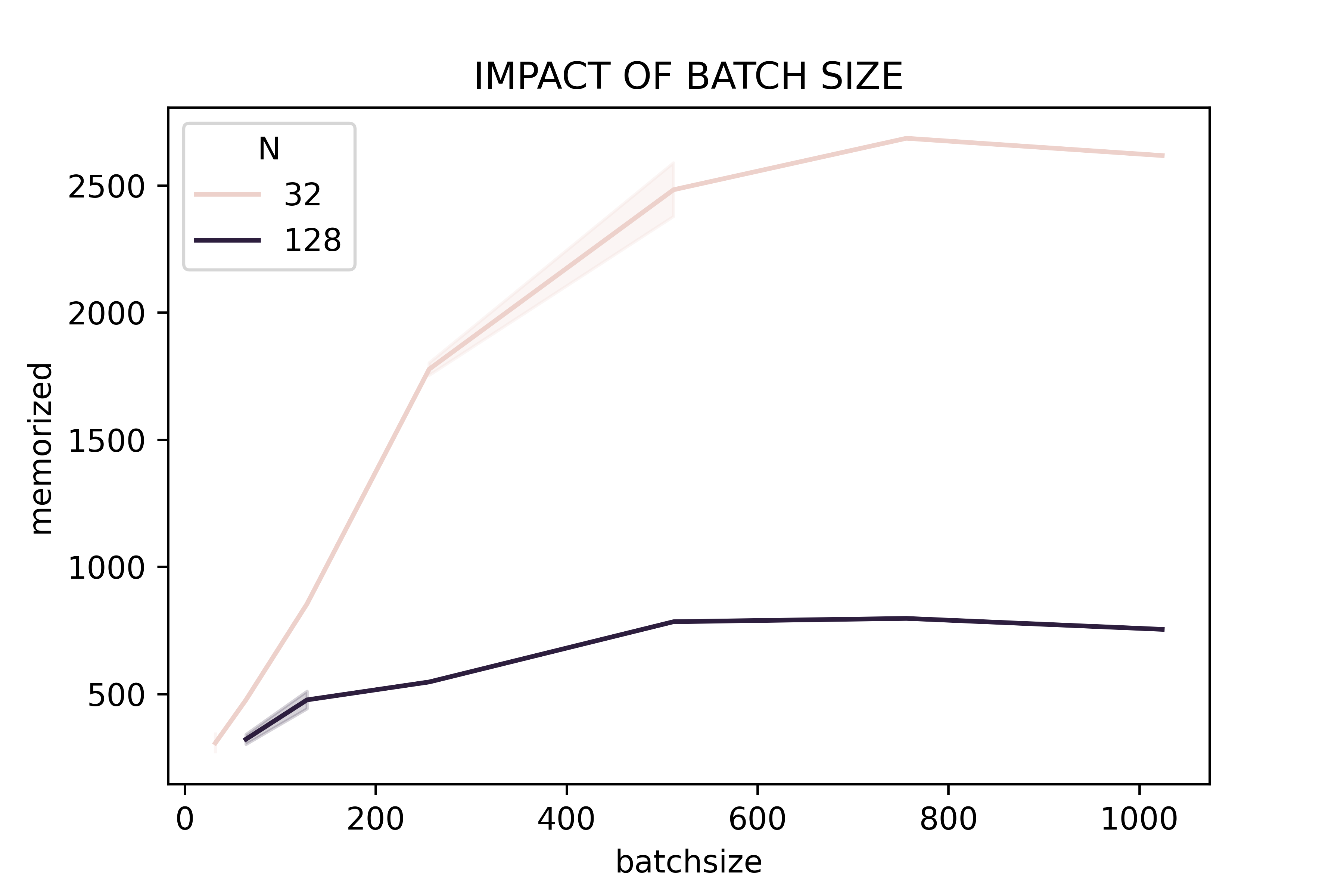}
    \caption{The impact of batch size on the performance for a model with $H=1$ and $B=512$}
    \label{fig:batchsize}
\end{figure}

The number of MAC memorized sequences of 32 or 128 tokens, as a function of batch size over a long training time, is shown in Fig. \ref{fig:batchsize}. The memorization is low with small batch sizes, which is also reflected in the large variation in the gradient noise. The capacity grows when the batch size increases and saturates, for the model sizes studied in this paper, after the batch size of 512.  

In this paper, the transformer model is based on the implementation in the popular x-transformers Python library \cite{wang_lucidrainsx-transformers_2024} and is trained using the Adam \cite{kingma_adam_2015} optimizer available in the PyTorch library, for which we use default settings. The selection of the optimal batch size can be estimated based on noise gradients; see, e.g., \cite{mccandlish_empirical_2018}. In this paper, we use a batch size of 512. In total, we trained approximately 500 models, the total training time was 260 hours on an Nvidia A100-SXM4-40GB GPU. Each training was restarted five times, and we selected the run with the highest number of memorized vectors. A run was terminated when there was no improvement in the number of memorized sequences after several epochs.    

\begin{figure}[t]
    \centering
    \includegraphics[width=7cm]{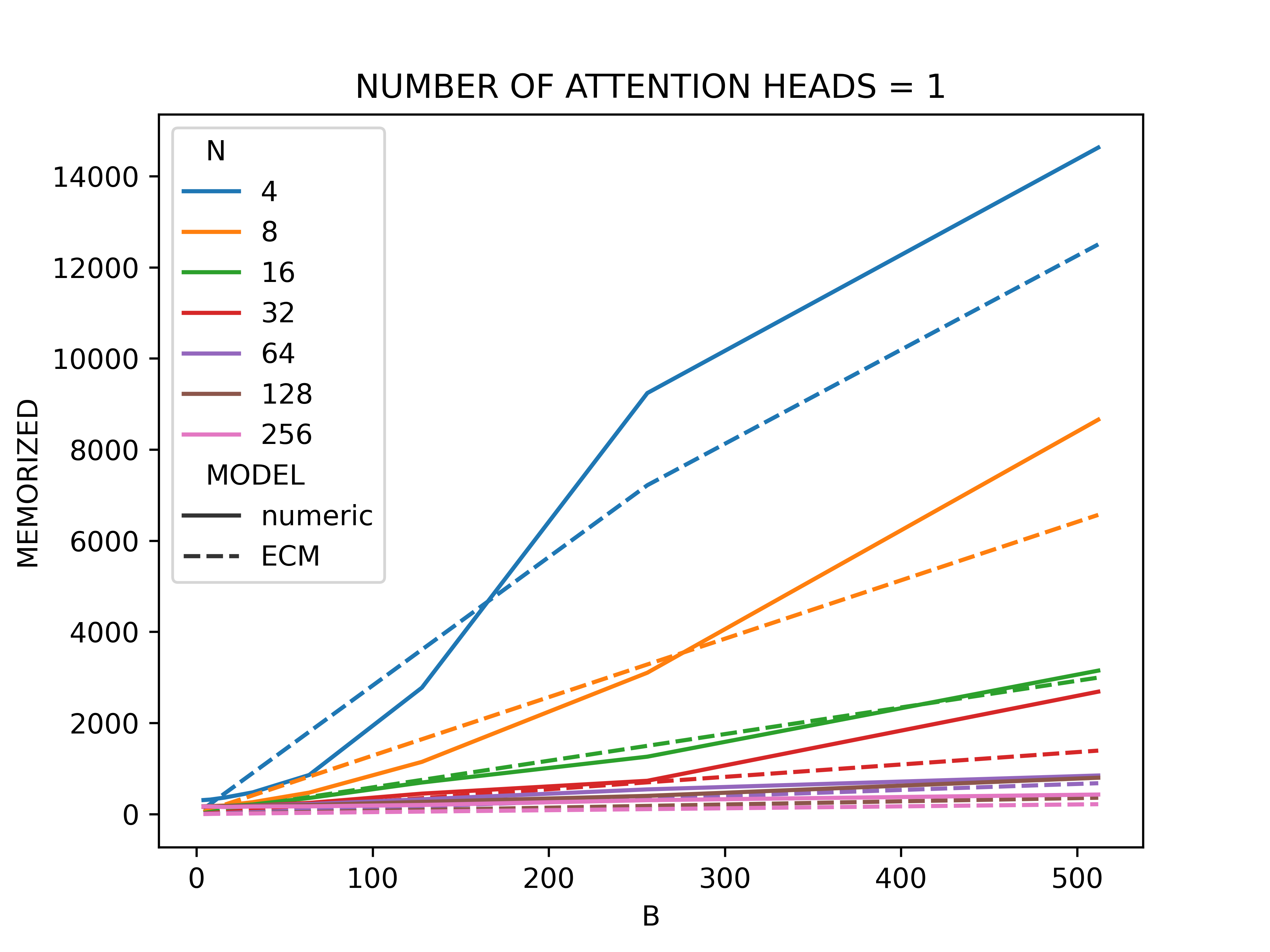}
    \vspace*{-4mm}
    \caption{Results of our ECM with $H=1$. Solid lines represent obtained capacity values whereas dashed lines represent our model predictions.}
    \label{fig:poly5}
\end{figure}

Fig. \ref{fig:poly5} shows the number of sequences fully memorized for MAC experiments in a one-layer network of $H=1$ attention heads, for different vector sizes $B$. The experiments reported in this paper were performed using a library size of 16000 sequences. Furthermore, the core model uses absolute positional embedding for the tokens and the dimension of the attention head is $d_h=128$, which is a common choice in several published models. 

\subsection{Detailed figures of measured model capacities}
\begin{figure}[t]
    \centering
    \includegraphics[width=6cm]{figs/our_model_H1.png}
    \includegraphics[width=6cm]{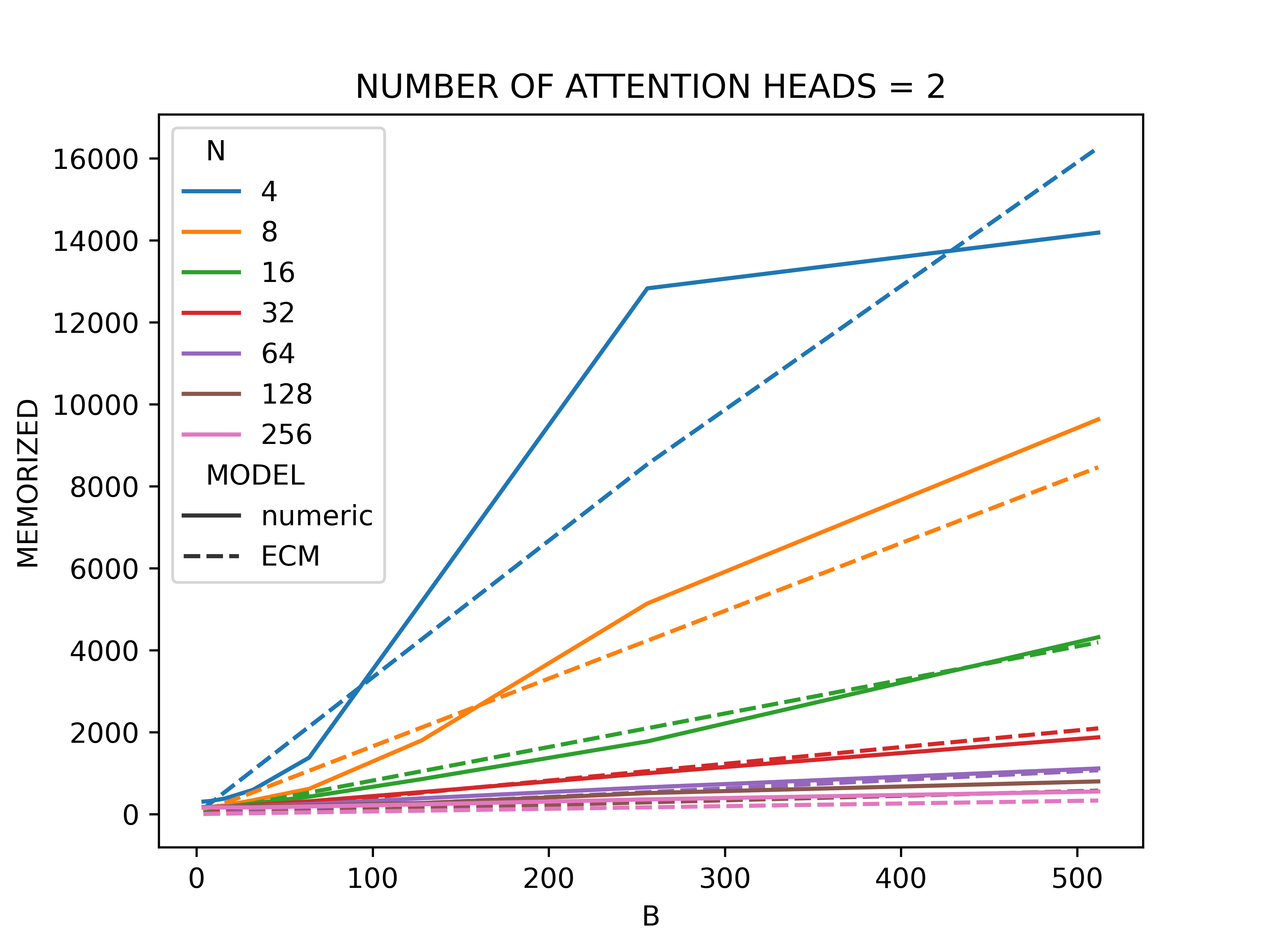}
    \includegraphics[width=6cm]{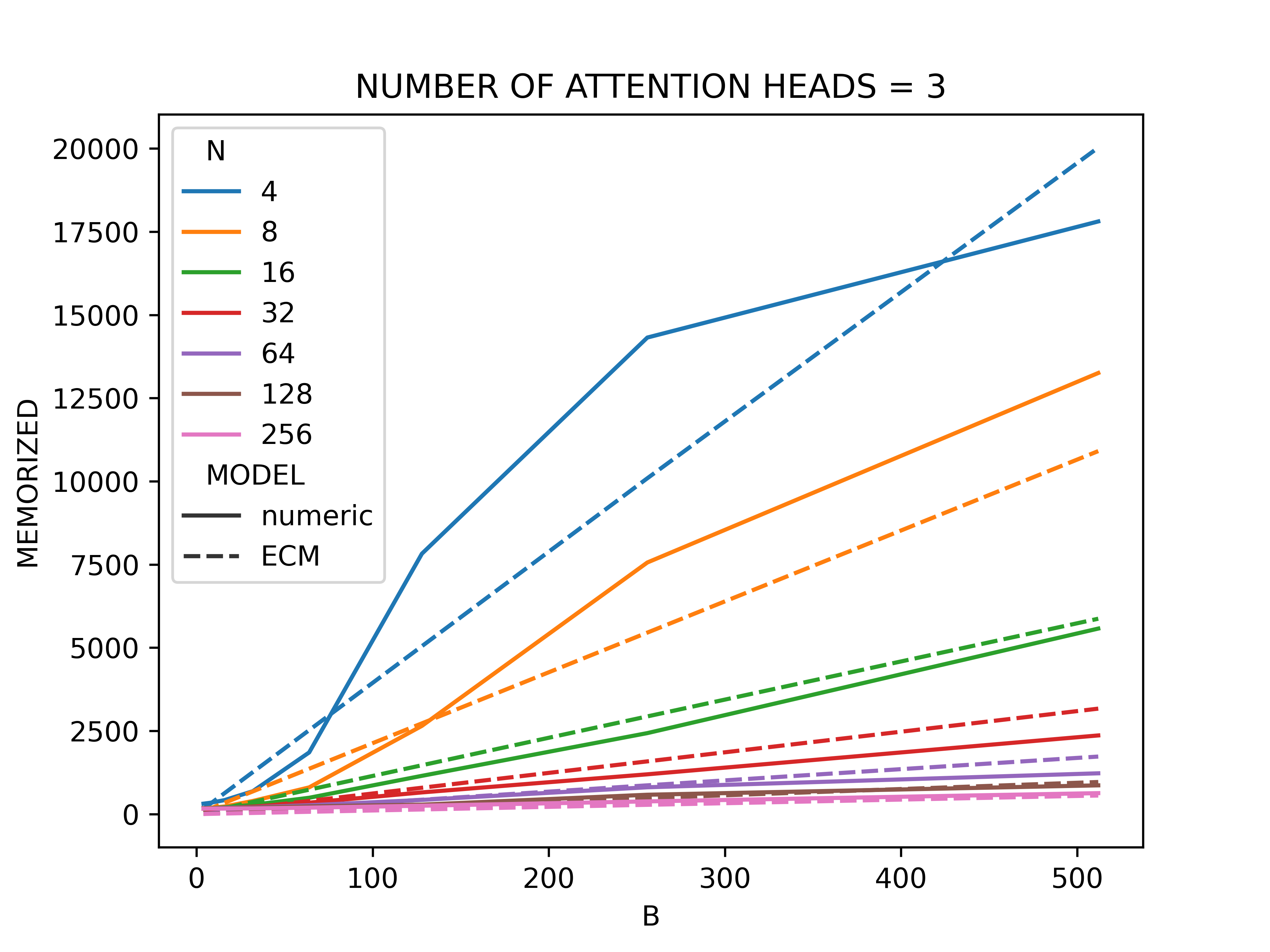}
    \includegraphics[width=6cm]{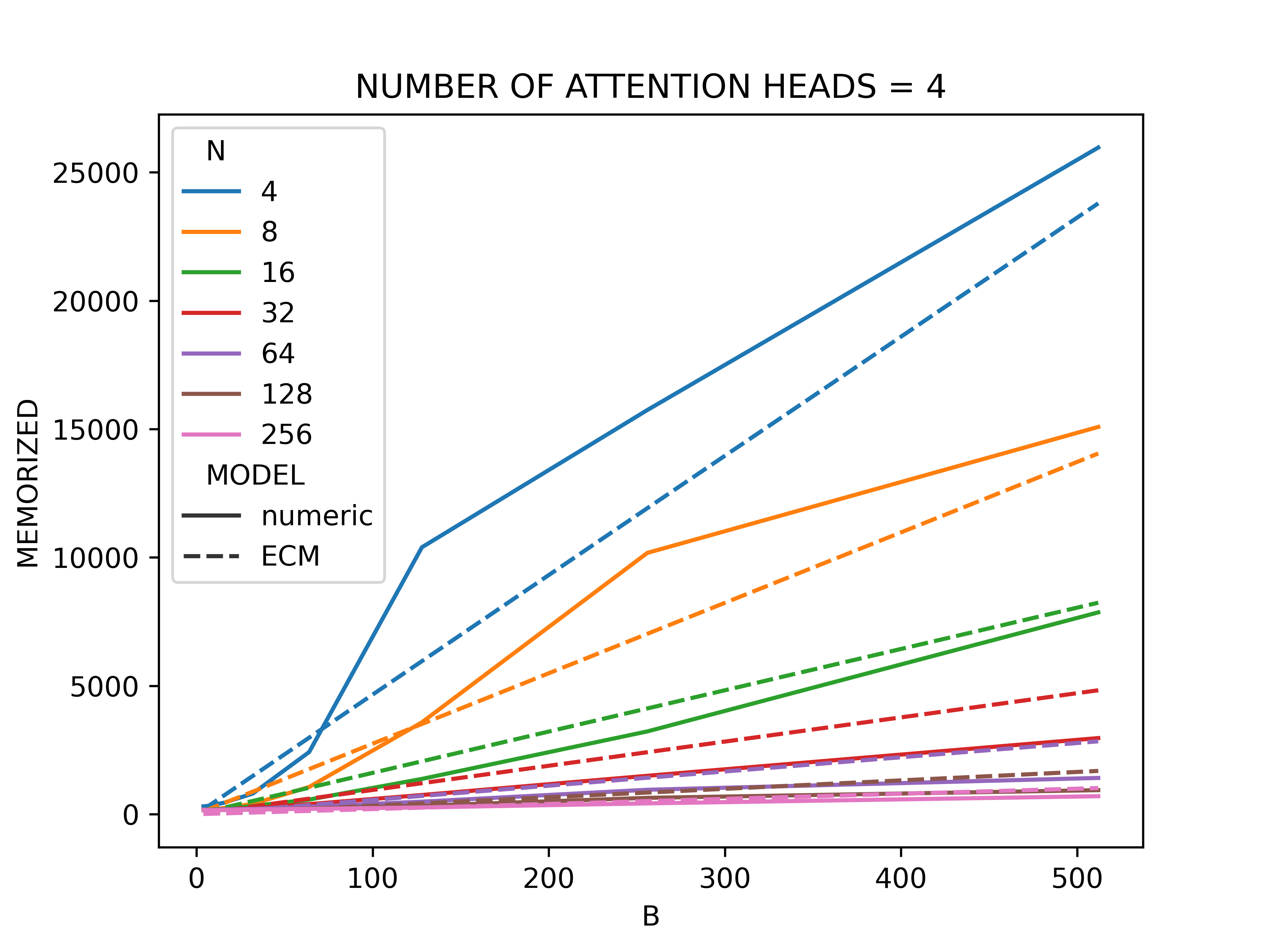}
    \caption{Results of our ECM. Solid lines represent obtained capacity values whereas dashed lines represent our model predictions. }
    \label{fig:poly5_app}
\end{figure}

The measured and predicted capacity figures for different transformer model architectures are shown in Figs. \ref{fig:poly5_app}.

\section{Empirical capacity model}
We utilize the results obtained from the experiments to develop an empirical capacity model for self-attention transformers. The model describes the relationship between the number of memorized sequences as a function of the transformer's hyperparameters. We build our model sequentially, that is, by decomposing it into intuitions obtained from examining the marginal effects of each of its inputs. These effects are then modeled using low-order algebraic functions. This allows for a capacity model that is interpretable and low in the number of parameters. We show that even such a simple formulation outperforms more complex, high-order polynomial models. What follows is a description of the marginal impacts of the model's inputs and how they are leveraged to derive a capacity function.

\subsection{Marginal impacts of hyperparamaters}
\begin{figure}[t]
    \centering
    \includegraphics[width=8cm]{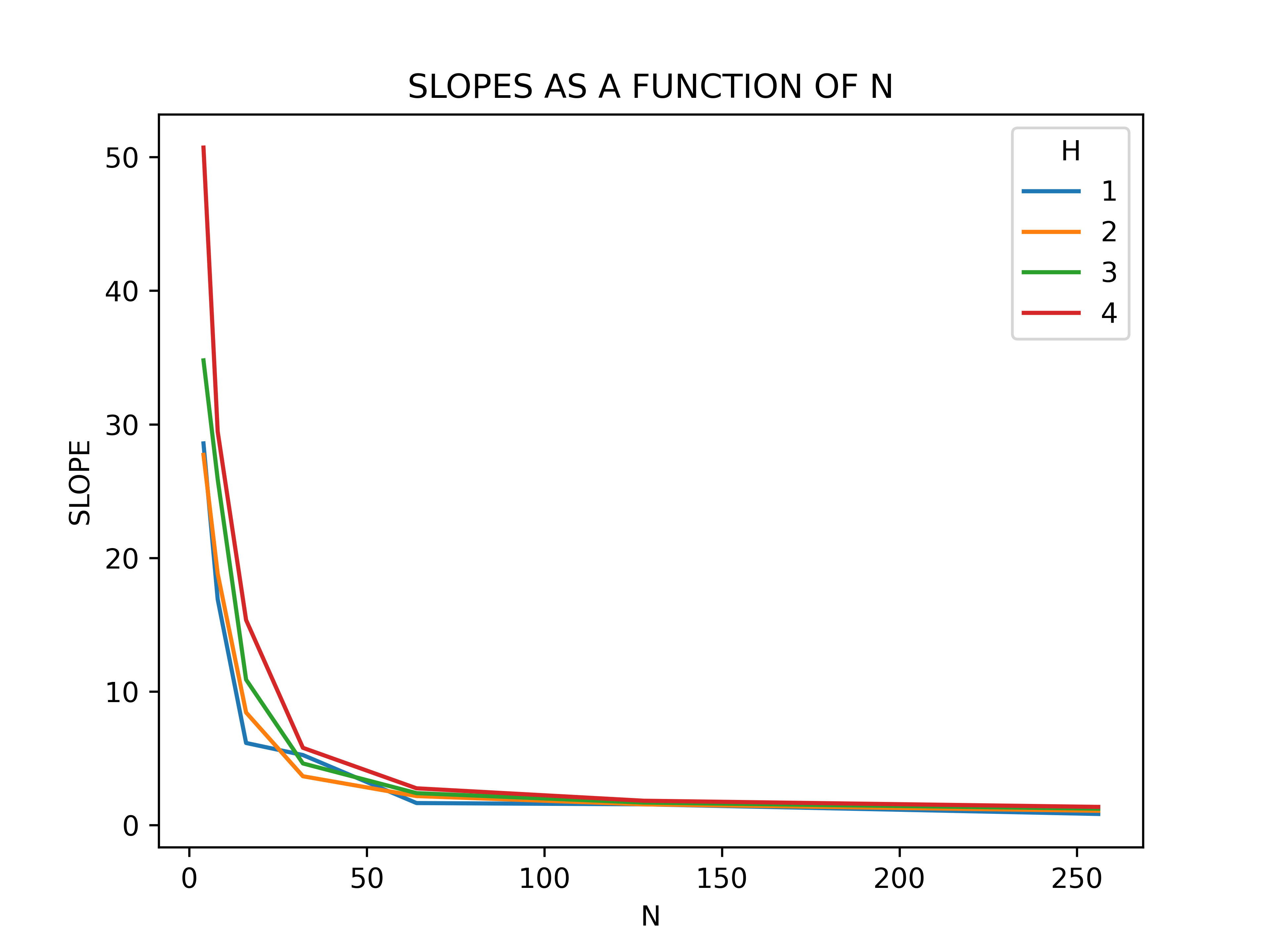}
    \caption{Slopes corresponding to the linear rise in $C$ as a function of $B$ with respect to $H$ and $N$.}
    \label{fig:slopes}
\end{figure}

We begin by capturing the marginal impact of $B$, that is, its behavior independent of the values of $H$ and $N$. This is shown by the solid lines of Fig.~\ref{fig:poly5}, which plots the number of memorized sequences as a function of $B$ at differing $H$ and $N$.
We notice that this number increases monotonically with larger $B$ values until a saturation point is reached. 
In our experiments, this saturation occurs at or slightly below the total number of sequences. 
We reflect this intuition in our modeling. Specifically, we first model the initial rise using a linear function.
A logistic function was also considered to capture the curves' slight sigmoidal nature; however, we found that the increase in the parameters in such a modeling was not empirically justified by a closer fit. 
In modeling a linear rise, it is necessary to determine the value of two parameters, the slope and the intercept. The intercept is trivial since a value of $B = 0$ implies a transformer with dimensionless features and thus no memorized sequences, resulting in an intercept of 0. The slope is determined by the values of $H$ and $N$, which we model by the function $f(H,N)$. We can thus model the presaturation of memorized sequences $C_{pre}$ as $C_{pre} = f(H,N)*B$.
 To model the slope values at the corresponding $H$ and $N$, $f(H,N)$, we first calculated them as the slope from the origin to the value at the highest $B$ before saturation. This then allows us to model the impact of $H$ and $N$ on $C$ as captured in Fig. \ref{fig:slopes}. In this regard, we notice an exponential decay in the slope with increasing $N$ values. Indeed, during our modeling, the exponential decay function was our first consideration. We found, however, that it does not capture the tapering that occurs at higher $N$ values. As such, we opted for a generalized rational function with a power function denominator. This allows us to better capture the changes in slope, whilst still maintaining an explainable formulation with few parameters. Finally, we notice that $H$ influences the rate of decay in the slope with the rate linearly rising at an increasing number of heads. This phenomenon corresponds to the exponent of the power term and can be modeled accordingly. Specifically, we raise $N$ to a linear function of $H$, resulting in the following slope function:    
 
\vspace*{-2mm}
\begin{equation}
    f(N,H) = \dfrac{a}{N^{b*H + c} + d} + e, 
\end{equation}
where $a$, $b$, $c$, $d$, and $e$ are parameters learned from our experiments. Their values are summarized in Table \ref{tab:parameters}.
Our slope formulation lends itself to a simple explanation for both input variables. $N$ represents the exponential decay in the rate at which sequences are memorized as a function of $B$, and $H$ influences the rate of this decay with higher values of $H$, resulting in a less pronounced drop.

\setlength{\tabcolsep}{5pt}
\begin{table}
\scriptsize
\caption{Learned parameters of our capacity models (slope and saturation).}
\centering
\begin{tabular}{l | c c c c c | c c} 
Layers & a & b & c & d & e & $\alpha$ & $\beta$ \\
\toprule
 1 & 145.27& -0.13 & 1.29 & 0.13 & 0.20 & 3762.70 & 8741.00 \\
 2 & 2.45 & -0.002 & 0.02 & -0.99 & -29.08 & 4413.10 & 14787.00 \\ 
\bottomrule
\end{tabular}
\label{tab:parameters}
\end{table}

\subsection{Empirical capacity model formulation}

Having assessed the marginal impacts of the hyperparameters, the Empirical Capacity Model can be formulated. Presaturation $C$, which is modeled by a linear function, increases without bound with respect to $B$ and thus does not model the point of saturation. This saturation point is a function of $H$, with the number of memorized sequences saturating at higher values with more attention heads. This rise is modeled by the linear function $\alpha*H + \beta$ which we found to offer a good fit with minimal additional parameters. The learned parameters of this saturation function on our data are shown also in Table \ref{tab:parameters}. This allows us to formulate our Empirical Capacity Model as the synthesis of the pre- and post-saturation states as follows:
\begin{equation}
    C = max(f(H,N)*B, \alpha * H + \beta) \label{eq:ecm}
\end{equation}
The fit of this model to our data is depicted graphically in Fig. \ref{fig:poly5}.
We can compare the fit of our model against a fifth-order polynomial of $H$, $N$, and $B$ as shown in the bottom row of this figure. We notice that despite the lower number of parameters (our model's seven compared to the polynomial's 56), our model provides a better fit. This is reflected in the mean absolute percentage errors (MAPE) of the two models. Specifically, our model has a MAPE of 0.497 compared to 1.0785 of the fifth-order polynomial. We analyze our proposed model in the following section.

\section{Analysis of the capacity model}
\begin{figure}
    \centering
    \includegraphics[width=7cm]{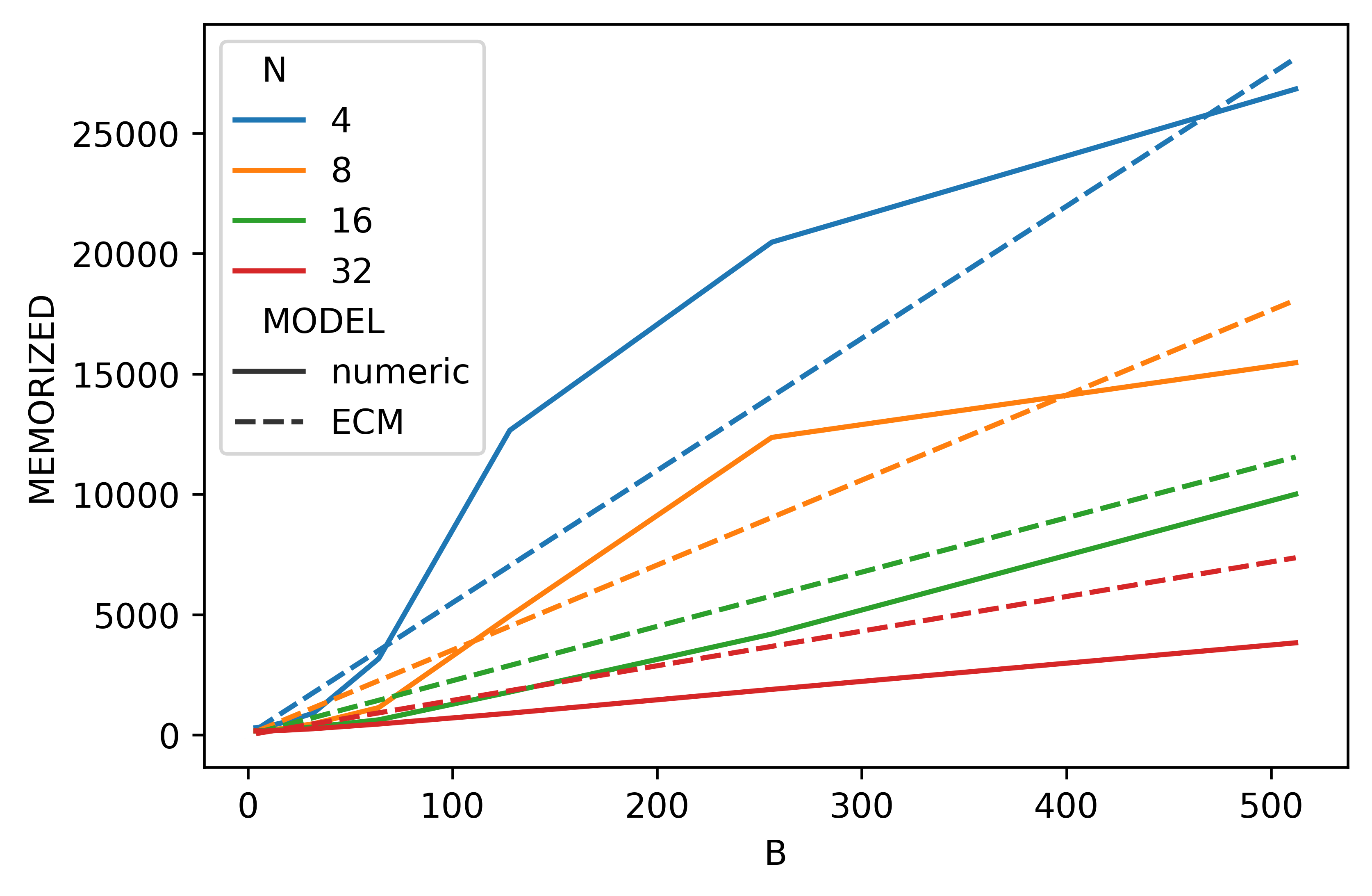}
    \vspace*{-4mm}    
    \caption{Real capacity measures (solid) and ECM predictions (dashed) for $H=5$.}
    \label{fig:H5_case}
\end{figure}

The parametric model of Equation \ref{eq:ecm} was optimized based on experiments with a library size of $K=16000$. The ECM was fitted with a selection of parameter ranges, and it is reasonably accurate inside the region. However, extrapolation outside of the range often leads to an under / overestimation of the measured capacity; see, for example, Fig.\ref{fig:H5_case} for $H=5$. 

\begin{figure}
    \centering
    \includegraphics[width=7cm]{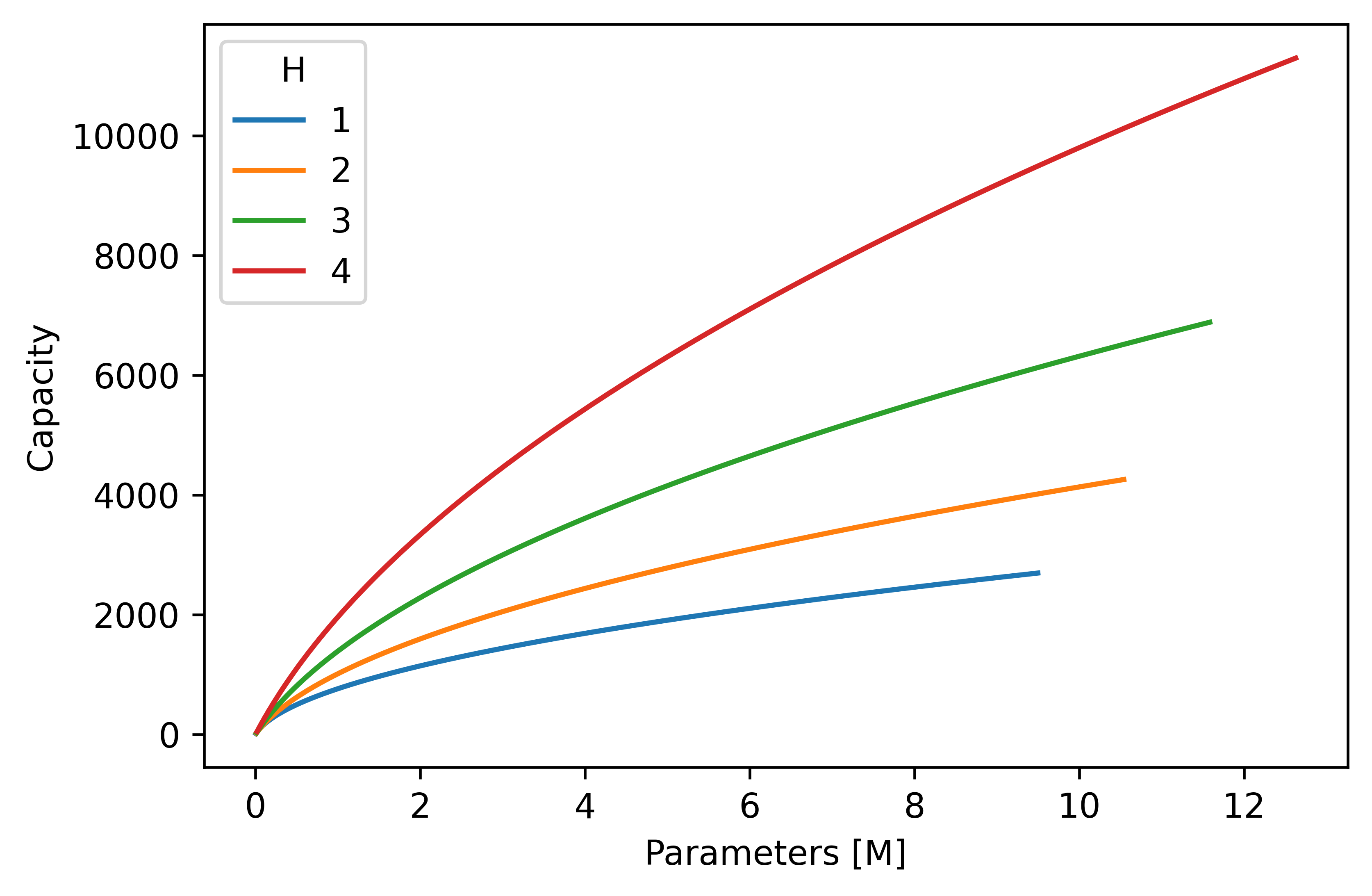}
    \vspace*{-4mm}    
    \caption{The model memory capacity as a function of the total number of trainable coefficients.}
    \label{fig:size-capacity}
\end{figure}

The ECM formula can be used to compare different architecture choices. For example, Fig. \ref{fig:size-capacity} shows the total capacity of the model as a function of the total number of coefficients for the sequences of length $N=64$. The plot shows that the capability increases significantly with more attention heads, $H=1\cdots 4$. For example, the capacity of a four-head model with 2M parameters is twice as high as that of a two-head model, that is, 1600 and 3300 fully memorized sequences, respectively. This agrees with the $O(HN)$ rule of \cite{mahdavi_memorization_2024}. 

We can leverage the ECM to examine the cost of adding additional parameters. In order to do so, we first define a measure that relates the number of parameters to the capacity of the model. Specifically, we define the memory per parameter (MPP) as the number of memorized tokens for each parameter in the model:
\begin{equation}
    MPP = \dfrac{max(f(H,N)*B, \alpha * H + \beta)}{\# parameters}
\end{equation}
By fixing the value of a hyperparameter and calculating the average MPP across the remaining hyperparameters, we can better understand the cost of adding additional parameters to the model. We notice that smaller networks use their parameters more efficiently as per the MPP than do larger ones. Furthermore, increasing the number of heads results in a lesser drop in efficiency than increasing $N$. As such, the head parameters are less computationally costly than those of $N$. 


\section{Discussion and conclusions}
In this paper, we evaluated the capacity of self-attention networks through an empirical analysis of their memorization potential in different settings. Such an analysis departs from the existing literature focusing on theoretical big-O bounds through the use of empirically derived capacities from experiments with synthetic data. In performing this work, our aim was to close the gap between the theoretical storage capacity and the factual, observed capacity of the transformer models that have garnered widespread interest and use in recent years.
To this end, we developed an empirical capacity model which can be used to infer the capacity of a network given its hyperparameters. This allows practitioners to have a principled tool for a priori hyperparameter selection that ensures the maximal utility of each parameter is achieved. This model was developed through an analysis of the marginal impacts of hyperparameters on network capacity. Specifically, we tested models on a synthetic autoregressive task in which the transformer must predict a subsequent token given a previously observed token sequence. The number of correctly predicted sequences was used as a proxy measurement of the capacity of the model. 
We ran models to capture the network capacities at different points in the hyperparameter space. The results of these models served as the basis for a model constructed to reflect insights about the behavior of each hyperparameter. Specifically, insights were synthesized to obtain the model as described in Equation \ref{eq:ecm}. This model, due to its simplicity and low number of parameters, is intuitive and outperforms more complex, higher-order polynomial models. 

In the future, we plan to expand our work to include more comprehensive and realistic data and testing. Specifically, we will obtain denser results that span a larger hyperparameter range and introduce the effects of the number of network layers. Furthermore, we will expand the data used to include natural language content. This will ensure that our model's conclusions are relevant outside of synthetic experiments and can be applied in real-world contexts. Finally, we will present a more thorough analysis of our capacity model and include guidelines for a priori hyperparameter selection. 

\clearpage
\bibliography{LargeLanguageModels,local_bibtex}
\bibliographystyle{splncs04}


\end{document}